\documentclass{article}

\PassOptionsToPackage{numbers, compress}{natbib}

     \usepackage[final]{neurips_2019}

\usepackage[utf8]{inputenc} 
\usepackage[T1]{fontenc}    
\usepackage{hyperref}       
\usepackage{url}            
\usepackage{booktabs}       
\usepackage{amsfonts}       
\usepackage{nicefrac}       
\usepackage{microtype}      
\usepackage[pdftex]{graphicx}
\usepackage{bm}
\usepackage{amsmath}
\usepackage{footnote}

\title{Using LSTMs for climate change assessment studies on droughts and floods}

\author{%
  Frederik Kratzert\thanks{LIT AI Lab \& Institute for Machine Learning, Johannes Kepler University Linz, Austria} \\
  \texttt{kratzert@ml.jku.at} \\
  \And
  Daniel Klotz\textsuperscript{$\ast$} \\
  \texttt{klotz@ml.jku.at} \\
  \And
  Johannes Brandstetter\textsuperscript{$\ast$} \\
  \texttt{brandstetter@ml.jku.at} \\
  \And
  Pieter-Jan Hoedt\textsuperscript{$\ast$} \\
  \texttt{hoedt@ml.jku.at} \\
  \And
  Grey Nearing\thanks{Department of Geological Sciences, University of Alabama, Tuscaloosa, AL United States} \\
  \texttt{gsnearing@ua.edu} \\
  \And
  Sepp Hochreiter\textsuperscript{$\ast$} \\
  \texttt{hochreit@ml.jku.at} \\
}

\begin{document}
\maketitle

\begin{abstract}
  Climate change affects occurrences of floods and droughts worldwide.
  However, predicting climate impacts over individual watersheds is difficult, primarily because accurate hydrological forecasts require models that are calibrated to past data.  
  In this work we present a large-scale LSTM-based modeling approach that - by training on large data sets - learns a diversity of hydrological behaviors. Previous work shows that this model is more accurate than current state-of-the-art models, even when the LSTM-based approach operates out-of-sample and the latter in-sample. In this work, we show how this model can assess the sensitivity of the underlying systems with regard to extreme (high and low) flows in individual watersheds over the continental US.
\end{abstract}

\section{Introduction}
Floods and droughts affect more people than any other type of weather-related natural hazard \citep{cred2015human}, and the propensities for both are likely to increase under climate change \citep{field2012managing, trenberth2014global}.

The most common strategy for assessing impacts of climate change on hydrologic systems uses models calibrated in individual catchments\footnote[1]{A \textit{catchment} (or \textit{basin}, \textit{watershed}) is the upstream area that drains to a certain point in a river.} against historical records \citep{vaze2010climate}. This strategy neglects the fact that a change in climate also leads to changes in the catchment characteristics, which is not realistic under climate change and other anthropogenic influences \citep{milly2008stationarity}. 
Currently, the primary challenges 
are: (i) simplistic models  \citep{vaze2010climate, clark2016characterizing}, (ii) unreliable parameter fitting  \citep{kirchner2006getting}, (iii) drastic performance degradation in large-scale (e.g. continental or global) modeling, \citep{archfield2015accelerating} and (iv) not accounting for changing environmental conditions in the setup \citep{merz2011time}.


Recently \citet{kratzert2019benchmarking} proposed an approach for large-scale hydrological simulation that outperforms a wide range of traditional models. It consists of an Long Short-Term Memory network (LSTM) \citep{hochreiter1997long} with a modified input gate, trained on meteorological time series data from hundreds of riverine systems, where static catchment characteristics are used to condition the model for a specific site. These characteristics comprise of topographic attributes (e.g. mean elevation, drainage area), soil properties (e.g. percentage of clay, soil conductivity), as well as climate and vegetation indices (e.g. mean annual precipitation, aridity, leaf area index). 
Furthermore, in previous publications \citet{kratzert2018agu} showed that the LSTM learns to model real hydrological processes (e.g., the amount of snow in a basin) in it's memory cell states without training on any type of direct snow-related data (except total precipitation).
This modeling of real hydrological processes provides at least some confidence that the LSTM learns some of the underlying physical process instead of just a simplistic mapping, e.g.\ on basis of spurious correlations.

In other words, there exists a proof-of-concept that deep learning can transfer information about hydrologic processes and behaviors between basins, time and unobserved locations. This is revolutionary in the Hydrological Sciences, where the problem of \textit{Prediction in Ungauged Basins} was the decadal problem of the International Association of Hydrological Sciences from 2003-2012 \citep{sivapalan2003iahs}, and is generally considered to be unsolved \citep{bloschl2016predictions}.

Here we use the EA-LSTM \citep{kratzert2019benchmarking} model to investigate which watersheds in the continuous USA have the largest sensitivities to climate-related forcings in extreme low-flow and high-flow periods.

\section{Methods}

\subsection{Entity-Aware Long Short-Term Memory Network (EA-LSTM)}

The EA-LSTM \citep{kratzert2019benchmarking} consists of an adapted LSTM cell, where static ($\bm{x}_s$) and dynamic input input features ($\bm{x}_d$) are used explicitly for different purposes: 
\begin{align}
    \bm{i} &= \sigma(\bm{W}_i\bm{x}_s + \bm{b}_i) \label{eq-ealstm-inputgate}\\
    \bm{f}[t] &= \sigma(\bm{W}_f\bm{x}_d[t] + \bm{U}_f\bm{h}[t-1] + \bm{b}_f) \label{eq-ealstm-forgetgate}\\
    \bm{g}[t] &= \mathrm{tanh}(\bm{W}_g\bm{x}_d[t] + \bm{U}_g\bm{h}[t-1] + \bm{b}_g) \label{eq-ealstm-cellinput}\\ 
    \bm{o}[t] &= \sigma(\bm{W}_o\bm{x}_d[t] + \bm{U}_o\bm{h}[t-1] + \bm{b}_o) \label{eq-ealstm-outputgate}\\
    \bm{c}[t] &= \bm{f}[t] \odot \bm{c}[t-1] + \bm{i} \odot \bm{g}[t] \label{eq-ealstm-cellstate}\\
    \bm{h}[t] &= \bm{o}[t] \odot \mathrm{tanh}(\bm{c}[t]),\label{eq-ealstm-hidden}
\end{align}
Here $t$ is the time step ($1 \leq t \leq T)$, $\bm{i}[t]$, $\bm{f}[t]$ and $\bm{o}[t]$ are the input gate, forget gate, and output gate, respectively, $\bm{g}[t]$ is the cell input, $\bm{h}[t-1]$ is the recurrent input, $\bm{c}[t-1]$ the cell state from the previous time step and $\bm{W}$, $\bm{U}$ and $\bm{b}$ the learnable parameters of the network. 

The EA-LSTM uses static input features $\bm{x}_s$ (observable catchment characteristics and climate indexes) to control the input gate. The dynamic input features $\bm{x}_d$ (meteorological time series data) are used in all other parts of the LSTM cell, together with the recurrent input $\bm{h}$. This setup allows the LSTM to activate different parts of the network for different basins, but also for similarly behaving basins to share certain parts of the network. 

\subsection{Data and Model}

To assess watershed susceptibility to climate-related risks in the continuous USA, we used pre-trained models published by \citet{kratzert2019benchmarking}. These models were trained on the data from 531 basins of the freely available CAMELS data set \citep{newman2014large, addor2017large}. The models predict daily streamflow using inputs that include five meteorological features (precipitation, min/max temperature, radiation, vapor pressure), 18 static catchment attributes, and 9 static climate indexes. 

\subsection{Assessing Climate Sensitivity}

To investigate which catchment characteristics influence droughts and floods, we used the method of \citet{morris1991factorial} to measure sensitivity of predicted streamflow to different input features during low-flow and high-flow periods. Low- and high-flow periods were defined below the $5^{th}$ percentile of the discharge distribution and above the $95^{th}$ percentile, respectively and act here as a proxy.
Specifically, we calculated the gradients of simulated streamflow w.r.t. $\bm{x}_s$ at each day of the simulation, and averaged the absolute gradients separately for each static input feature (catchment characteristics and climate indexes) over the low- and high-flow periods. Averaged values were normalized to [0,1] separately in each basin \citep{saltelli2004sensitivity}, so that the features could be ranked according to their relative influence. 

\section{Results and Discussions}

\begin{figure}
  \centering
  \includegraphics[width=\linewidth]{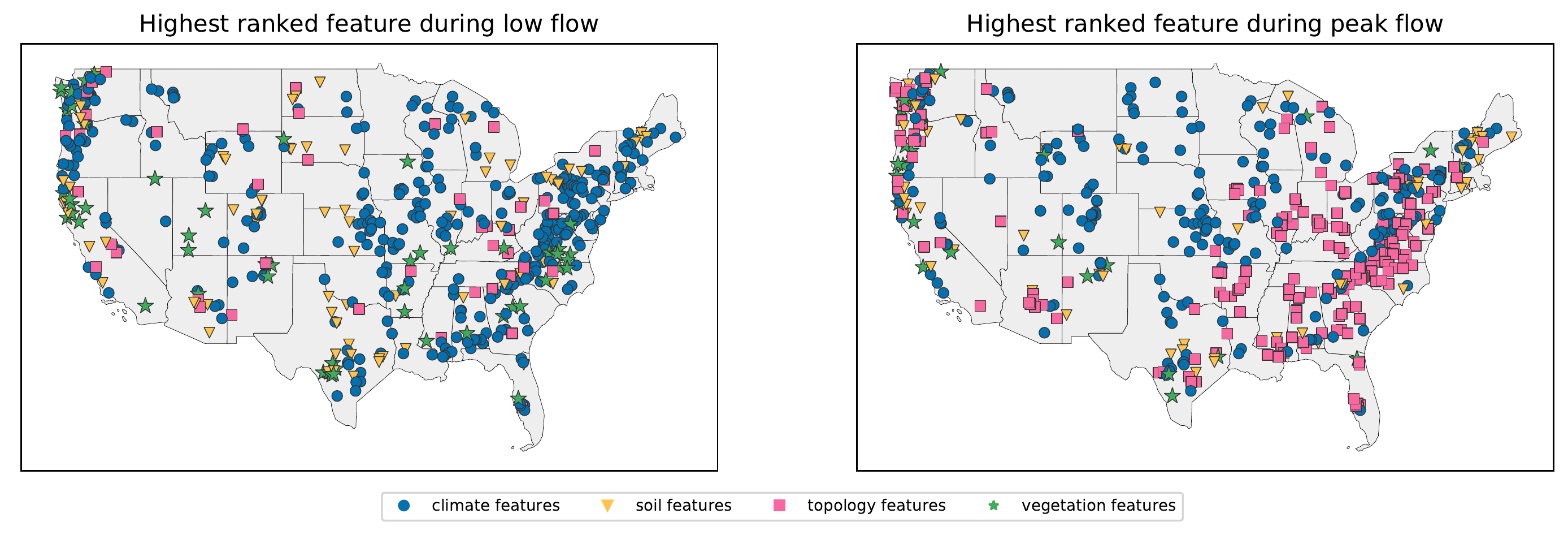}
  \caption{Highest ranked feature for low flow periods (as proxy for droughts) on the left-hand side and floods on the right hand-side. The features were grouped into either climate- (blue circles), soil- (yellow triangle), topology- (pink square) or vegetation-type (green asterisk) feature.}
  \label{fig_basins}
\end{figure}

Averaged over all basins, the top 5 features for the low flow periods are (1) mean annual precipitation sum, (2) aridity, (3) duration and (4) frequency of high precipitation events, (5) frequency of low precipitation events and for peak flow periods (1) drainage area, (2) mean annual precipitation, (3) mean elevation, (4) aridity, (5) high precipitation duration.

To investigate their respective spatial patterns individual $\bm{x}_{s,i}$ were grouped into categories related to: climate, soil, topography, and vegetation. Figure \ref{fig_basins} shows the spatial distribution of the most sensitive feature groups during low-flow on the left, and high-flow periods on the right. There are three important takeaways from this figure. First, climate features are more dominant during low-flow, while topology features are more dominant during high-flow. Second, there is clear geographical clustering, especially for high-flow periods where climate features are typically the most sensitive in the central part of the continent (Rocky Mountains, Great Plains, Central Plains), in southern California, and on the eastern Appalachian foothills. This is largely due to the strong influence of the aridity feature in these relatively dry basins. Rivers that have highest ranking soil and vegetation features appear dispersed over the data. Third, clear (and realistic) differences in the model sensitivity are visible over the continent. This indicates that the model - at least potentially - provides useful information for local water managers to assess climate-related risks in individual watersheds.

In summary, the results suggest that droughts (or low-flow periods) are more sensitive to changes in climate. However, we caution that this study is only a proof-of-concept showing that this kind of sensitivity analysis might be useful for climate change impact assessment. Future work will focus on using the EA-LSTM (or a modified version) to run counterfactual scenario analyses and using it in conjunction with other methods and verification tests to assess potential impacts.

\section{Outlook and Future Work}

Hydrological modeling usually assumes that the catchment characteristics of the environmental systems are stationary over long periods of time \citep{vaze2010climate,hall2014understanding,clark2016characterizing}. However, this "stationarity is dead" \citep{milly2008stationarity} and hydrologists have struggled to build models that are regionally applicable, and yet accurate in individual basins - the EA-LSTM is more accurate than existing models, even for basins that were not used for training \citep{kratzert2019benchmarking,kratzert2019pub}. By learning simultaneously from a large number of basins under different eco-hydrological regimes, the EA-LSTM can assess influences of different types of boundary conditions, and has the potential to adapt to changing hydrologic or climatic conditions. 

Currently, the used basin and climate characteristics are derived once for the entire data period. However the model structure allows for dynamic input features (e.g., dynamic climate and vegetation indexes, or dynamic anthropogenic demand indexes). Feeding the model with evolving input features, e.g. as obtained from climate projections, could make it possible to account for changes to individual basins by building on experience that is learned from modeling the diverse training data set. This opens the door to fundamentally new possibilities for large-scale hydrological impact assessment under climate change, that is able to maintain its local relevance.

\bibliographystyle{abbrvnat}
\bibliography{bib-refs}

\end{document}